\documentclass{ifacconf}
\usepackage{amsmath,amssymb,amsfonts}
\let\theoremstyle\relax 
\usepackage{amsthm}
\usepackage{mathrsfs}
\usepackage{algorithm}
\usepackage{algpseudocode}
\usepackage{graphicx}      
\usepackage{natbib}        
\usepackage{xcolor}
\newcommand{\ipP}[2]{\left\langle #1,#2 \right\rangle_{P}} 
\begin{document}
\begin{frontmatter}

\title{Lyapunov Neural ODE State-Feedback Policies} 

\thanks[footnoteinfo]{This work was supported by the National Science Foundation under Grants 2130734 and 2317629.}

\author[First]{Joshua Hang Sai Ip}
\author[First]{Georgios Makrygiorgos}
\author[First]{Ali Mesbah}

\address[First]{Department of Chemical and Biomolecular Engineering,
University of California, Berkeley, CA 94720, USA \\
\{ipjoshua, gmakr, mesbah\}@berkeley.edu}



\begin{abstract}Deep neural networks are increasingly used as an effective parameterization of control policies in various learning-based control paradigms.  
For continuous-time optimal control problems (OCPs), which are central to many decision-making tasks, control policy learning can be cast as a neural ordinary differential equation (NODE) problem wherein state and control constraints are naturally accommodated. 
This paper presents a NODE approach to solving continuous-time OCPs for the case of stabilizing a known constrained nonlinear system around a target state. The approach, termed Lyapunov-NODE control (L-NODEC), uses a novel Lyapunov loss formulation that incorporates an exponentially-stabilizing control Lyapunov function to learn a state-feedback neural control policy, bridging the gap of solving continuous-time OCPs via NODEs with stability guarantees. The proposed Lyapunov loss allows L-NODEC to guarantee exponential stability of the controlled system, as well as its adversarial robustness to perturbations to the initial state. 
The performance of L-NODEC is illustrated in two problems, including a dose delivery problem in plasma medicine. In both cases, L-NODEC effectively stabilizes the controlled system around the target state despite perturbations to the initial state and reduces the inference time necessary to reach the target. 
\end{abstract}


\end{frontmatter}

\section{Introduction}
\label{sec:introduction}
Optimal control is foundational to decision-making for complex dynamical systems \citep{athans2007optimal,lewis2012optimal,stengel1994optimal}. Efficient solution methods for optimal control problems (OCPs) are essential for tasks such as optimization-based parameter and state estimation, optimal experimental design, and model-based control \citep{bryson2018applied}. Solving continuous-time OCPs is challenging, especially for systems with nonlinear dynamics and path constraints, since these OCPs
involve infinitely many decision variables in the form of time-varying functions. Various techniques are developed to solve continuous-time OCPs, including direct methods that approximate the original infinite-dimensional problem as a finite-dimensional one via discretization of the time-varying functions  \citep{teo1991unified,biegler2002advances},
\textcolor{black}{though discretization can yield a large number of decision variables. Alternatively, 
indirect methods solve the necessary optimality conditions using Pontryagin’s maximum principle or the Hamilton–Jacobi–Bellman equation \citep{hartl1995survey,luus2019iterative}, but may lack scalability to higher-dimensional problems. 
There are also global optimization methods (e.g., \cite{chachuat2006global,rodrigues2022efficient}), which fall beyond the scope of this work.}

In this paper, we adopt a learning perspective to solving continuous-time OCPs. In particular, (deep) neural networks (NN) are widely used to represent control policies in reinforcement learning (RL) for Markov decision processes \citep{sutton2018reinforcement}, which fundamentally relies on approximately solving an optimal control problem \citep{bertsekas2019reinforcement}. Additionally, NN control policies have received growing attention in differentiable control (e.g., \cite{jin2020pontryagin,drgovna2022differentiable}) and imitation learning for predictive control (e.g., \cite{mesbah2022fusion,paulson2020approximate}). The interest in NN control policies stems from their scalability for high-dimensional problems and representation capacity due to universal approximation theorem \citep{barron1993universal}. However, learning NN policies can be sample inefficient, which is especially a challenge in applications where the policy must be learned via interactions with a real system. On the other hand, when a system model is available in the form of differential equations, the model can be used to formulate a continuous-time OCP while a NN policy is used to parameterize the time-varying function of decisions as a state-feedback control policy \citep{sandoval2023neural}. This allows for approximating the otherwise intractable OCP, while incorporating path and terminal state constraints into an OCP; what remains a largely open problem in RL.

\textcolor{black}{The latter approach to solving continuous-time OCPs with a NN control policy follows the same strategy as learning neural ordinary differential equations (NODEs) \citep{chen2018neural}. NODEs comprise a class of NN models that replace the discrete hidden layers in dense NNs with a parameterized ODE that represents continuous-depth models, describing temporal evolution of the hidden states in dynamic inference. Such an interpretation of dynamical systems as a learnable function class offers distinct benefits for time-series modeling (e.g., \cite{rahman2022neural,linot2023stabilized}). In continuous-time OCPs, system dynamics can be viewed as a composition of a known ODE model and a NN control policy embedded in the dynamics, forming a NODE structure. 
This setting is an instance of the universal differential equation framework \citep{rackauckas2020universal}, which embodies the idea of using various types of NNs within physics-based models.}  
The advantages of \emph{neural} control policies resulting from the NODE structure over traditional NN policies include: (i) allowing the use of modern numerical ODE solvers  that leverage adaptive step sizes based on desired accuracy and speed while ensuring numerical stability, which is essential for inference of stiff systems; (ii) handling states and inputs over arbitrary (sampling) time intervals, alleviating the need to discretize data on fixed time intervals; and (iii) memory efficiency of NODEs due to the use of the adjoint method \citep{givoli2021tutorial} for gradient computations in the backward pass, circumventing backpropagation through the numerical solver. The latter eliminates the need to store intermediate values in the forward pass, reducing the memory footprint in learning control policies. Another advantage of using NODEs for control lies in the ability to perform system identification and control policy design in a unified framework \citep{bachhuber2023neural,chi2024nodec}, enabling performance-oriented model learning \citep{gevers2005identification,makrygiorgos2022performance}.
The dynamic nature of NODEs naturally lends itself to leveraging control-theoretic tools to provide desirable structures such as stability in learning neural control policies for OCPs. \textcolor{black}{Stability is particularly crucial in optimal control for establishing robustness of the optimal solution to ensure the controlled system reaches its desired state despite perturbations.
}

For unconstrained autonomous systems, \citet{rodriguez2022lyanet} proposed a method for learning NODEs using the concept of exponentially-stabilizing control Lyapunov functions (CLFs) \citep{ames2014rapidly}. The main idea of this method is to use the supervised loss of NODE as a potential function, so that the NODE training loss embeds both the learnable dynamics and the potential function. Inspired by this notion, we present a new approach to solving continuous-time OCPs, termed Lyapunov-NODE control (L-NODEC), for constrained nonlinear systems with known dynamics. L-NODEC seeks to learn a state-feedback neural control policy that achieves exponential convergence to a target state. To this end, a novel Lyapunov loss formulation is presented that embeds an exponentially-stabilizing CLF to guarantee exponential stability of the controlled system. Additionally, we prove that L-NODEC  guarantees adversarial robustness to uncertain initial states by deriving an upper bound on the deviation of the terminal state from the target state. 
The superior performance of L-NODEC over NODEC with no stability guarantees is demonstrated using two case studies, including an OCP application in plasma medicine.

\section{Preliminaries}
\label{sec:problem_statement}

\subsection{Problem Formulation}

We consider the problem of designing a state-feedback neural control policy $\pi(x(t))$ that solves the following OCP in the Mayer form 
\begin{subequations}
\label{eq:ocp}
\begin{align}
    \min_{\theta} \quad & \mathbb{E}_{x_0 \sim P_{x_0}}[\Phi(x(T))] \label{eq:objective}\\
    \text{subject to:} \quad & \dot{x}(t) = f(x(t), t) + h(x(t), t)\pi_\theta(x(t)), \label{eq:dynamics}\\
    & g(x(t), \pi_\theta(x(t))) \leq 0, \quad \quad \ \ \, \forall t \in [0,T], \label{eq:path_constraints}\\
    & \pi_\theta(x(t)) \in \mathcal{U} = [u^{\text{lb}}, u^{\text{ub}}], \quad \forall t \in [0,T] , \label{eq:control_bounds} 
\end{align}
\end{subequations}
where $x \in \mathbb{R}^{n_{x}}$ is the state in the time interval $[0, T]$, where $T$ denotes the final time; $P_{x_0}$ denotes a known distribution with a bounded support of admissible initial states $x_0 = x(0)$; $f: \mathbb{R}^{n_x} \times \mathbb{R} \rightarrow \mathbb{R}^{n_x}$ and $h: \mathbb{R}^{n_x} \times \mathbb{R} \rightarrow \mathbb{R}^{n_x \times n_u}$ represent the known drift and input matrix fields, respectively; $g: \mathbb{R}^{n_x} \times \mathcal{U} \rightarrow \mathbb{R}^{n_g}$ represents state and input constraints; $\pi_\theta: \mathbb{R}^{n_x} \to \mathcal{U} \subset \mathbb{R}^{n_u},
$ is a static state-feedback control policy parameterized by $\theta \in \mathbb{R}^{n_\theta}$; $\Phi(x(T))$ denotes the terminal cost; and $u^{\text{lb}}, u^{\text{ub}} \in \mathbb{R}^{n_u}$ denote the lower and upper bounds on the control input, respectively.  
The cost $\Phi: \mathbb{R}^{n_x} \rightarrow \mathbb{R}_+$ is defined as the $P$-weighted squared distance of the current state to the target state $z$
\begin{equation}
\label{eq:terminal_cost}
\Phi(x) = (x - z)^T P (x - z) = \|x - z\|_P^2,
\end{equation}
where $P \in \mathbb{R}^{n_x \times n_x}$ is a user-specified positive definite weight matrix ($P \succ 0$), and $\|\cdot\|_P$ denotes the $P$-weighted norm defined as $\|x\|_P := \sqrt{x^T P x}$. 
The choice of $P$ encodes the relative importance of different state components when measuring deviation from the target.

Since the OCP in \eqref{eq:ocp} specifies only a terminal cost, it provides merely a boundary condition, leaving significant freedom in the trajectory taken by the system between the initial and target states. Thus, to encourage trajectories to reach the target state more rapidly while ensuring stability, we reformulate the terminal cost \eqref{eq:terminal_cost} into a Lyapunov-based function to embed a guarantee of exponential stability in the closed-loop system.  
This formulation is particularly relevant for continuous-time OCP applications where the control objective is to drive the state to a desired target with a guaranteed convergence rate despite uncertainty in the initial state. 
Furthermore, by enforcing exponential convergence over $t \in [0,T]$, the proposed approach enables early termination once the system reaches the target, reducing extra control effort and computational burden.

\subsection{Neural Ordinary Differential Equations}

NODEs provide a useful framework for learning ODEs of the form~\eqref{eq:dynamics}. \textcolor{black}{In this work, we define NODEs that admit an affine structure $\mathcal{F}_\theta(x, t)$ as} 
\begin{equation}
\label{NODE}
\frac{{dx}}{{dt}} 
= \mathcal{F}_\theta(x, t) := f(x, t) + h(x, t)\pi_\theta(x),
\end{equation}
where $\pi_\theta(x)$ is obtained from solving \eqref{eq:ocp}. NODEs are related to the well-known ResNet \citep{haber2017stable}. The hidden layers in a ResNet architecture can be viewed as discrete-time Euler's integration
of \eqref{NODE}; that is, ResNet can be thought of as learning discrete-time dynamics with a fixed time step. Given $\theta$, \eqref{NODE} can be numerically integrated over a desired time interval for inference of system dynamics. Backpropagation for learning NODEs can be efficiently implemented via the adjoint method \citep{givoli2021tutorial}. The procedure involves solving a ``backward-in-time'' ODE associated with \eqref{NODE}, known as the adjoint ODE. Solving the adjoint ODE yields gradients of the loss function with respect to states $x$ at each time step. These gradients can then be utilized to calculate the loss function gradients with respect to the learnable parameters $\theta$ using the chain rule \citep{chen2018neural}. 
The NODE framework enables the use of numerical ODE solvers with adaptive time-steps, which is especially useful for inference of stiff dynamical systems. Additionally, the ODE solver embedded in NODEs allows for systematic error growth control and trading off numerical accuracy with efficiency. Despite these advantages, the standard NODE framework does not impose desired properties, such as stability and robustness to uncertain initial conditions, within the learned dynamics $\mathcal{F}_\theta(x, t)$. This can potentially lead to unstable trajectories.

\subsection{Lyapunov Stability}
Inspired by NODEs, this work presents an approach for learning an exponentially stabilizing state-feedback neural control policy in \eqref{eq:ocp}.
Lyapunov theory generalizes the notion of stability of dynamical systems by reasoning about the convergence of a system to states that minimize a potential function \citep{Lyapunov1961}. Potential functions are a special case of dynamic projection. 

\color{black}
\textbf{Definition 1 (Dynamic projection \citep{taylor2019control})}.
A continuously differentiable function $V : \mathcal{X} \rightarrow \mathbb{R}$ is a dynamic projection if there exists a state $z \in \mathcal{X}$ and constants $\underline{\sigma}$, $\bar{\sigma} > 0$ that satisfy\footnote{Definition 1 holds for any norm, but $l_2$ norm is adopted for defining dynamic projection in this work.}
\begin{equation}
\label{dynamic_projection}
\forall x \in \mathcal{X} : \underline{\sigma} \|x - z\|_2^2 \leq V(x) \leq \bar{\sigma} \|x - z\|_2^2.
\end{equation}
\color{black}
The notion of dynamic projection can be used to define the exponential stability of $\eqref{NODE}$ as follows. 

\textbf{Definition 2 (Exponential stability).}
The NODE system \eqref{NODE} is exponentially stable if there exists a positive-definite dynamic projection potential function $V$ and a constant $\kappa > 0$ such that all solution trajectories of \eqref{NODE} for all $t \in \textcolor{black}{[0, T]}$ satisfy
\begin{equation}
\label{exponential_stability}
V(x(t)) \leq V(x(\textcolor{black}{0}))e^{-\kappa t}.
\end{equation}
We use an exponentially stabilizing control Lyapunov function (ES-CLF) to ensure exponential stability of \eqref{NODE}. 

{\color{black}
\textbf{Theorem 1 (Exponentially stabilizing control Lyapunov function \citep{ames2014rapidly}).}
\textit{
For the NODE system \eqref{NODE}, a locally continuously differentiable positive-definite dynamic projection potential function $V$ is an ES-CLF for an initial state $x_0$ if there exist $\kappa > 0$ and a locally Lipschitz continuous policy $\pi_\theta(x)$ that satisfy
}
\begin{equation}
\label{ES-CLF_min}
    \underset{\theta \in \Theta}{\inf}\ \left[\frac{\partial V}{\partial x} \bigg|^\top_x \mathcal{F}_\theta(x, t) + \kappa V(x) 
    \right] \leq 0, \quad \forall t \in [0,T].
\end{equation}
\textit{
According to \eqref{ES-CLF_min}, there exists $\bar{\theta} \in \Theta$ such that
}
\begin{equation}
\label{ES-CLF}
  \frac{\partial V}{\partial x} \bigg|^\top_{x} \mathcal{F}_{\bar{\theta}}(x, t) + \kappa V(x) \leq 0,
\end{equation}
\textit{which implies \eqref{NODE} parameterized by $\bar{\theta}$ is exponentially stable with respect to the potential function $V$.}

Inequality \eqref{ES-CLF} enforces a contraction condition on $V(t)$, termed local invariance \citep{rodriguez2022lyanet}.

\section{Lyapunov-NODE Control (L-NODEC)}
\label{sec:theory}

This section presents the proposed L-NODEC approach for learning an exponentially stabilizing state-feedback neural control policy $\pi_{\theta}(x)$. 
For the OCP~\eqref{eq:ocp}, the potential function is defined as
\begin{equation}
\label{potential_function}
V(x(t)) := \Phi(x(t)) = (x(t) - z)^\top P (x(t) - z).
\end{equation}

Given $P \succ 0$, we have
\begin{equation}
\label{wmin_wmax}
\begin{split}
\lambda_{\min}\,\|x(t)-z\|_2^2 & \leq (x(t) - z)^\top P (x(t) - z) \\
&\leq \lambda_{\max}\,\|x(t)-z\|_2^2,
\end{split}
\end{equation}
where $\lambda_{\min},\lambda_{\max} > 0$ correspond to the smallest and largest eigenvalues of $P$, respectively. 
Equation~\eqref{wmin_wmax} implies $V(x(t))$ is a dynamic projection per Definition 1.

We utilize \eqref{ES-CLF} to define a pointwise Lyapunov loss $\mathcal{V}(x(t))$ with respect to the states of the controlled system \eqref{NODE}. The pointwise Lyapunov loss is defined as violation of the local invariance for the dynamic projection $V(x(t))$. That is,  
\begin{equation}
\label{pointwise_loss}
\mathcal{V}(x(t)) = \max \left\{ 0,\frac{\partial V}{\partial x}\bigg|^\top_{x} \mathcal{F}_\theta(x, t) + \kappa V(x(t)) \right\}.
\end{equation}
Note that the pointwise Lyapunov loss will take on a non-zero value when it violates the local invariance. 
To obtain the proposed Lyapunov loss, we integrate the pointwise Lyapunov loss \eqref{pointwise_loss} in time and take the expectation over the distribution of the initial state $x_0 \sim P_{x_0}$
\begin{equation}
\label{lyapunov_loss}
\mathscr{L}(\theta) = \mathbb{E}_{x_0\sim P_{x_0}} \left[\int_{0}^{T} \mathcal{V}(x(t))\ dt\right].
\end{equation}
The Lyapunov loss~\eqref{lyapunov_loss} corresponds to the expected violation of the local invariance for the entire time $[0, T]$ over all initial conditions $x_0 \sim P_{x_0}$. 
The L-NODEC approach uses the Lyapunov loss \eqref{lyapunov_loss} to learn $\pi_\theta$, ensuring that the resulting policy yields exponentially stable state trajectories for the controlled system \eqref{NODE}. The rationale behind using the Lyapunov loss \eqref{lyapunov_loss} lies in  the underlying structure of the pointwise loss function \eqref{pointwise_loss}, which enforces the non-negativity condition $\mathcal{V} \geq 0$. According to Theorem 1, this condition is equivalent to satisfying the exponential stability condition \eqref{ES-CLF}, as the controlled system is penalized with non-negative loss when $V$ is not an ES-CLF. 

\textbf{Theorem 2 (Exponential stability of L-NODEC).}
If there exists $\theta^* \in \Theta$ such that $\mathscr{L}(\theta^*)=0$, then $V$ in \eqref{potential_function} is an ES-CLF (i.e., satisfies \eqref{ES-CLF}). Consequently, the controlled system \eqref{NODE} with $\pi_{\theta^*}$ satisfies
\[
\|x(t)-z\|_P \;\le\; e^{-\kappa t/2}\,\|x_0-z\|_P
\quad \forall \, t\in[0,T],\; x_0\sim P_{x_0}.
\]
\begin{proof}
We first consider a single initial state drawn from $P_{x_0}$. From $\mathscr{L}(\theta^*)=0$ and the definition of $\mathcal{V}$, we have $\frac{\partial V}{\partial x}^\top \mathcal{F}_{\theta^*}(x,t)+\kappa V(x)\le 0$ for all $t\in[0,T]$. 
With $V(x)=\|x-z\|_P^2$, this gives $\dot V(t)\le -\kappa V(t)$, which satisfies the conditions for ES-CLF, as specified in Theorem 1. This also implies the definition of exponential stability $\eqref{exponential_stability}$ is satisfied. Thus, taking square roots gives the stated bound
$\|x(t)-z\|_P \le e^{-\kappa t/2}\|x_0-z\|_P$. Due to the non-negative integral in \eqref{lyapunov_loss}, the same proof holds for all $x_0 \sim P_{x_0}$ without loss of generality.
\end{proof}

Furthermore, L-NODEC yields neural control policies that are adversarially robust to perturbations in the initial state $x(0)$. This property arises from the local invariance of \eqref{pointwise_loss} since exponential stability ensures the states converge exponentially to the target $z$. To analyze adversarial robustness, we define stable inference dynamics. 



\textbf{Definition 3 ($\delta$-stable inference dynamics).} 
For the system \eqref{NODE} with the optimal control policy $\pi_{\theta^*}(x)$ 
and potential function $V(x(t))$ in \eqref{potential_function}, 
the pair $(x_0, z)$ where $x_0 \sim P_{x_0}$ has $\delta$-stable inference dynamics for $\delta > 0$ if:
\begin{enumerate}
    \item \textbf{Exponential stability:} The potential function $V(x(t))$ satisfies the ES-CLF condition $\dot{V}(x(t)) \le -\kappa V(x(t))$ for some $\kappa > 0$;
    \item \textbf{$\delta$-final bound:} At the final inference time $t = T$, the potential satisfies 
    \[
    V(x(T)) \le e^{-\kappa T} V(x(0)) \le \delta.
    \]
\end{enumerate}

The state trajectories of \eqref{NODE} under the optimal policy $\pi_{\theta^*}(x)$ are exponentially stable, 
and the potential function at the final time $T$ is bounded by $\delta$. 
The smallest $\delta$ that can be guaranteed for a given initial state is
\[
\delta^* = e^{-\kappa T} V(x(0)).
\]

\textbf{Theorem 3 (Adversarial robustness to initial state perturbations).}
Given the potential function $V(x(t))$ in \eqref{potential_function}, assume there exist $\theta^*$ and $\kappa>0$ such that the ES-CLF \eqref{ES-CLF} holds for a state trajectory of \eqref{NODE} under $x_0$ and  $\pi_{\theta^*}$.
Let $x_\epsilon(0)=x_0+\epsilon$, where $x_0 \sim P_{x_0}$ and $\|\epsilon\|_\infty\le\bar\epsilon$, and $x_\epsilon(\cdot)$ be its corresponding trajectory under $\pi_{\theta^*}$.
Then, at the terminal time $T$,
\begin{equation}
\label{eq:V_decay_ic_main_T}
V\big(x_\epsilon(T)\big)
\;\le\; e^{-\kappa T}\,V\big(x_\epsilon(0)\big)
\;=\; e^{-\kappa T}\,\|x(0)+\epsilon - z\|_P^2,
\end{equation}
and
\begin{equation}
\label{eq:delta_bound_P_T}
V\big(x_\epsilon(T)\big)
\;\le\; e^{-\kappa T}\Big(\,\|x(0)-z\|_P
      + \sqrt{\lambda_{\max}(P)}\,\sqrt{n_x}\,\bar\epsilon\,\Big)^2.
\end{equation}


\emph{Proof.}
\emph{Step 1 (Exponential decay along the perturbed trajectory).}
Along $x_\epsilon(\cdot)$, we have
\begin{equation}
\label{eq:Vdot_inequality_eps}
\frac{d}{dt}V\!\big(x_\epsilon(t)\big)
= \frac{\partial V}{\partial x}\Big|_{x_\epsilon(t)}^{\!\top}
  \mathcal{F}_{\theta^*}\!\big(x_\epsilon(t),t\big)
\,\le\, -\kappa\,V\!\big(x_\epsilon(t)\big).
\end{equation}
Using the Grönwall inequality \citep{khalil2002nonlinear},
\begin{equation}
\label{eq:V_decay_eps_all_t}
V\big(x_\epsilon(t)\big)\,\le\, e^{-\kappa t}\,V\big(x_\epsilon(0)\big),
\quad t\in[0,T].
\end{equation}
Evaluating \eqref{eq:V_decay_eps_all_t} at $t=T$ gives \eqref{eq:V_decay_ic_main_T}.

\emph{Step 2 (Explicit terminal bound).}
From \eqref{eq:V_decay_ic_main_T},
\begin{equation}
\label{eq:terminal_P_start}
V\big(x_\epsilon(T)\big)
\,\le\, e^{-\kappa T}\,\|x(0)+\epsilon-z\|_P^2.
\end{equation}
By the triangle inequality in the $P$-norm,
\begin{equation}
\label{eq:triangle_P}
\|x(0)+\epsilon-z\|_P
\;\le\; \|x(0)-z\|_P + \|\epsilon\|_P.
\end{equation}
Therefore,
\begin{equation}
\label{eq:terminal_before_bound}
V\big(x_\epsilon(T)\big)
\;\le\; e^{-\kappa T}\Big(\,\|x(0)-z\|_P + \|\epsilon\|_P\,\Big)^2.
\end{equation}
Using the Rayleigh-quotient and $\ell_\infty\!\to\!\ell_2$ bounds,
\begin{equation}
\label{eq:epsilonP_bound}
\|\epsilon\|_P \;\le\; \sqrt{\lambda_{\max}(P)}\,\|\epsilon\|_2
\;\le\; \sqrt{\lambda_{\max}(P)}\,\sqrt{n_x}\,\bar\epsilon.
\end{equation}
Substituting \eqref{eq:epsilonP_bound} into \eqref{eq:terminal_before_bound} yields \eqref{eq:delta_bound_P_T}.
\hfill$\square$

Inequality \eqref{eq:V_decay_ic_main_T} provides the exact terminal decay, while \eqref{eq:delta_bound_P_T} makes this decay explicit in terms of $\|x(0)-z\|_P$, the number of states $n_x$, and $\bar\epsilon$. 
By \eqref{eq:V_decay_ic_main_T}, for any perturbed initial state $x_\epsilon(0)$, the pair $(x_\epsilon(0),z)$
is $\delta$-stable on $[0,T]$ with $\delta = e^{-\kappa T}V(x_\epsilon(0))$. 
In particular, any
\[
\delta \;\ge\; e^{-\kappa T}\Big(\,\|x(0)-z\|_P
      + \sqrt{\lambda_{\max}(P)}\,\sqrt{n_x}\,\bar\epsilon\,\Big)^2
\]
guarantees $\delta$-stable inference dynamics for all $\|\epsilon\|_\infty\le\bar\epsilon$.

 \textbf{Remark (Admissible optimal policies under the ES-CLF condition).}
Along the state trajectories of \eqref{NODE} under the optimal control policy $\pi_{\theta^*}(x)$, we have
\[
\dot V(x)
= \frac{\partial V}{\partial x}^\top \mathcal{F}_{\theta^*}(x,t)
= 2(x-z)^\top P\!\left[f(x,t)+h(x,t)\,\pi_{\theta^*}(x)\right].
\]
The ES-CLF inequality $\dot V(x)+\kappa V(x)\le 0$ yields
\begin{align}
2(x - z)^\top P f(x, t)
&\;+\; 2(x - z)^\top P h(x, t)\,\pi_{\theta^*}(x) \nonumber\\
&\;+\; \kappa (x - z)^\top P (x - z)
\;\le\; 0.
\end{align}
Isolating the policy term gives a linear inequality constraint on $\pi_{\theta^*}$, satisfying the Lyapunov decrease condition 
\begin{align}
(x - z)^\top P\,h(x, t)\,\pi_{\theta^*}(x)
&\;\le\;
-(x - z)^\top P f(x, t) \nonumber\\
&\quad
-\frac{\kappa}{2}\,(x - z)^\top P (x - z).
\label{eq:control_halfspace_standard}
\end{align}
Given the $P$-weighted inner product
$\ipP{a}{b}:=a^\top P b$ and norm $\|v\|_P^2:=v^\top P v$, the admissible set of optimal neural control policies is defined as
\begin{equation}
\label{eq:U_adm}
\begin{aligned}
\mathcal{U}_{\mathrm{adm}}(x,t)
:= \{ & \,\pi_{\theta^*}\in \mathcal{U}:\;
 \ipP{x-z}{h(x,t)\,u} \le {} 
 \\
& -\ipP{x-z}{f(x,t)} - \tfrac{\kappa}{2}\,\|x-z\|_P^2 \,\}.
\end{aligned}
\end{equation}
This result indicates that the admissibility of control policies is restricted by $P$, which sets the decay direction $P(x-z)$, the target $z$, the decay rate $\kappa$, and the known dynamics $f,h$. In~\eqref{eq:U_adm}, the term $\ipP{x-z}{f(x,t)}$ is the drift’s projection along the Lyapunov gradient direction; if positive, it tends to increase $V$. 
The control projection $\ipP{x-z}{h(x,t)\,\pi_\theta(x)}$ must be sufficiently negative to offset this drift and supply the margin $(\kappa/2)\,\|x-z\|_P^2$.

\section{L-NODEC Learning Framework}

In this section, we first describe how system constraints can be integrated into the L-NODEC approach. We then present the algorithm used to learn the neural control policy. Finally, we discuss the sampling procedure for approximating the expectation in the Lyapunov loss \eqref{lyapunov_loss}.

\subsection{System Constraints}

L-NODEC can be modified to enforce state and input constraints \eqref{eq:path_constraints} and \eqref{eq:control_bounds}, respectively. The inputs designed by the neural control policy can be constrained in the last layer of the neural policy by using a sigmoid activation function, defined as $\sigma(\cdot):\mathbb{R}\to[0,T]$, i.e.,
\begin{equation}
\label{input_constraint}
u_i \;=\; u^{\mathrm{lb}}_i \;+\; \big(u^{\mathrm{ub}}_i - u^{\mathrm{lb}}_i\big)\,\sigma(\cdot),
\qquad i\in\{1,\dots,n_u\}.
\end{equation}

To enforce state constraints \eqref{eq:path_constraints}, penalty terms in the form of the squared violation \citep{bertsekas1975} are appended to the pointwise Lyapunov loss \eqref{pointwise_loss}. This leads to
\begin{equation}
\label{eq:pointwise_loss_constrained_smooth}
\begin{aligned}
\mathcal{V}_c(x,t)
&= \max\!\left\{0,\;
\frac{\partial V}{\partial x}\bigg|_{x}^{\!\top}\,\mathcal{F}_\theta(x,t)
\;+\;\kappa\,V(x)\right\} \\
&\quad +\;
\beta\,\big\|\,\mathcal{S}_\tau\!\big(g(x,\pi_\theta(x))\big)\big\|_2^{\,2},
\end{aligned}
\end{equation}
where $
\mathcal{S}_\tau(\psi):=\tau\,\ln\!\big(1+\exp(\psi/\tau)\big)$ is the softplus function, with $\tau>0$ and
 penalty weight $\beta>0$. This is a popular approach to enforcing state constraints, as penalty convergence theorem guarantees a feasible solution to the reformulated unconstrained optimization problem, which is equivalent to solving a constrained optimization problem with Karush–Kuhn–Tucker multipliers \citep{chen2016,DiPillo1994}. However, since \eqref{eq:pointwise_loss_constrained_smooth} is composed of two terms, there exists a tradeoff between exponential stability of the controlled system \eqref{NODE} and satisfaction of state constraints \eqref{eq:path_constraints}, as further discussed in Section~\ref{sec:case_studies}.
 
\textbf{Remark.} Other approaches can also be used to enforce state constraints. Control barrier functions (CBFs) enforce forward invariance of a safe set by imposing a differential condition along the closed-loop trajectories; CBFs are well-suited for hard safety constraints, but can be conservative in practice \citep{cohen2020approximate}. Alternatively, the augmented Lagrangian method augments the objective with linear multipliers and quadratic penalties,
often yielding tighter satisfaction of constraints at increased computational cost \citep{bergounioux1997augemented}.

\subsection{Neural Control Policy Learning}
\label{subsec:training_adjoint}

To learn the neural control policy $\pi_\theta(x)$, we minimize the Lyapunov loss \eqref{lyapunov_loss} over $[0,T]$ using mini-batches of initial states sampled from $P_{x_0}$. To this end, we discretize $[0,T]$ by $t_i=i\Delta t$ with $\Delta t:=T/\Gamma$ and $i=0,\dots,\Gamma$. The discretized objective takes the form of
\begin{equation}
\label{lyapunov_loss_discretized}
\begin{aligned}
\mathscr{L}(\theta)
\;\approx\;
\mathbb{E}_{x_0\sim P_{x_0}}
\Big[
\sum_{i=0}^{\Gamma-1}&\mathcal{V}_c\!\big(x(t_i),t_i\big)\,\Delta t
\Big],
\end{aligned}
\end{equation}
where $\mathcal{V}_c$ is either the pointwise Lyapunov loss \eqref{pointwise_loss}, or its constrained version \eqref{eq:pointwise_loss_constrained_smooth}. At each training iteration, a mini-batch $\{x_0^{(b)}\}_{b=1}^{B}$ is drawn from $P_{x_0}$. For each $b$, we simulate the dynamics of the controlled system~\eqref{NODE}, i.e.,
\begin{equation}
\label{eq:closed_loop_dynamics}
\begin{aligned}
\dot x^{(b)}(t)
&= \mathcal{F}_\theta\!\big(x^{(b)}(t),t\big) \\
&= f\!\big(x^{(b)}(t),t\big)
 + h\!\big(x^{(b)}(t),t\big)\,\pi_\theta\!\big(x^{(b)}(t)\big)
\end{aligned}
\end{equation}
on $[0,T]$, and accumulate the per-trajectory loss
\begin{equation}
\label{lyapunov_loss_batch}
\begin{aligned}
\mathscr{L}^{(b)}(\theta)
&:= \sum_{i=0}^{\Gamma-1}\mathcal{V}_c\!\big(x^{(b)}(t_i),t_i\big)\,\Delta t \\
&\approx \int_0^T \ell\!\big(x^{(b)}(t),t;\theta\big)\,dt,
\qquad \ell(x,t;\theta):=\mathcal{V}_c(x,t).
\end{aligned}
\end{equation}
The batch loss is $\mathscr{L}_{\text{batch}}(\theta)=\tfrac{1}{B}\sum_{b=1}^{B}\mathscr{L}^{(b)}(\theta)$.

For each trajectory $x^{(b)}(\cdot)$, the \emph{adjoint} (or \emph{costate}) $a^{(b)}(t)\in\mathbb{R}^{n_x}$ is the Lagrange multiplier that enforces the dynamics in the variational formulation and measures how a perturbation of $x^{(b)}(t)$ affects the future loss. Here, the adjoint is integrated backward with terminal condition $a^{(b)}(T)=0$. For $b=1,\dots,B$, the adjoint ODE is
\begin{equation}
\label{adjoint_dynamics_correct}
\begin{aligned}
\frac{d a^{(b)}(t)}{dt}
&= -\left(\frac{\partial \mathcal{F}_\theta}{\partial x}\big(x^{(b)}(t),t\big)\right)^{\!\top} a^{(b)}(t) \\
&\quad - \frac{\partial \ell}{\partial x}\big(x^{(b)}(t),t;\theta\big),
\qquad a^{(b)}(T)=0.
\end{aligned}
\end{equation}
Hence, the gradient of the per-trajectory loss is
\begin{equation}
\label{adjoint_gradient_correct}
\begin{aligned}
\frac{d\mathscr{L}^{(b)}}{d\theta}
= \int_{0}^{T}
\Big[
\frac{\partial \ell}{\partial \theta}\big(x^{(b)}(t),t;\theta\big)
&+ a^{(b)}(t)^{\!\top}\frac{\partial \mathcal{F}_\theta}{\partial \theta}\big(x^{(b)}(t),t\big)
\Big]\,dt .
\end{aligned}
\end{equation}
The L-NODEC learning framework is summarized in Algorithm 1.



\begin{algorithm}[t!]
\caption{L-NODEC via adjoints (mini-batch initial state sampling)}
\label{algo_lnodec_training}
\begin{algorithmic}[1]
\State \textbf{Inputs:} $M$ (iterations), $B$ (batch size), $\Gamma$ (time steps), horizon $T$, step $\Delta t\!\leftarrow\!T/\Gamma$, learning rate $\alpha$, decay rate $\kappa$, penalty $\beta$, initial state distribution $P_{x_0}$
\State \textbf{Initialize:} neural control policy parameters $\theta$
\For{$k=1$ to $M$}
    \State Sample $\{x_0^{(b)}\}_{b=1}^{B}\sim P_{x_0}$; set $\mathscr{L}_{\text{batch}}=0$, $G_{\text{batch}}=0$
    \For{$b=1$ to $B$} \label{line:pertraj}
        \State \textit{Forward:} integrate \eqref{eq:closed_loop_dynamics} on $[0,T]$ with $\pi_\theta(x)$ and saturation \eqref{input_constraint}
        \State Accumulate $\mathscr{L}^{(b)}(\theta)\approx \sum_{i=0}^{\Gamma-1}\mathcal{V}_c(x^{(b)}(t_i),t_i)\,\Delta t$ using \eqref{eq:pointwise_loss_constrained_smooth}
        \State \textit{Backward:} integrate \eqref{adjoint_dynamics_correct} from $t=T$ to $0$ with $a^{(b)}(T)=0$
        \State Compute $G^{(b)}=\dfrac{d\mathscr{L}^{(b)}}{d\theta}$ via \eqref{adjoint_gradient_correct}; \quad
               $\mathscr{L}_{\text{batch}} \mathrel{+}= \mathscr{L}^{(b)}/B$, \;
               $G_{\text{batch}} \mathrel{+}= G^{(b)}/B$
    \EndFor
    \State \textit{Update:} $\theta \leftarrow \theta - \alpha\, G_{\text{batch}}$
\EndFor
\State \textbf{Output:} learned policy parameters $\theta^*$
\end{algorithmic}
\end{algorithm}  

\section{Case Studies}
\label{sec:case_studies}

L-NODEC is demonstrated on a benchmark double integrator problem and a cold atmospheric plasma system with prototypical applications in plasma medicine. The performance of L-NODEC is compared to that of neural ODE control (NODEC) \citep{sandoval2023neural}.\footnote{In both case studies, the neural control policy $\mathbb{\pi_{\theta}}$ is parameterized by 3 hidden layers of 32 nodes each, and the RMSProp optimizer is used for policy training. The code for the double integrator case study can be found at https://github.com/ipjoshua1483/LNODEC.}

\subsection{Double Integrator}

\begin{figure*}
    \centering
    \includegraphics[width=1\linewidth]{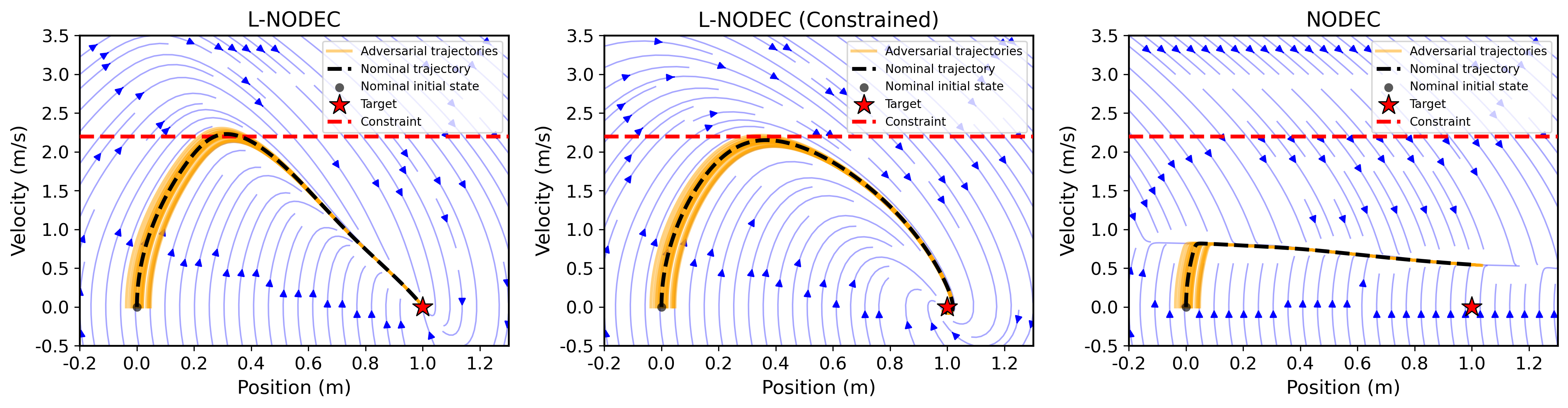}
    \caption{Phase portraits of the controlled double integrator system for L-NODEC (left), L-NODEC Constrained (middle), and NODEC (right). The nominal trajectory and adversarial trajectories are shown in black and orange, respectively. The initial states for the adversarial trajectories are based on different initial positions $x_1$ sampled from a uniform distribution over [-0.05, 0.05] with the nominal initial velocity 0. The streamlines in the vector field and the constraint for velocity $x_2$ are shown in blue and red, respectively.}
    \label{fig:double_integrator_phase_portraits}
\end{figure*}

The setup follows \citep{logsdon1992}
\begin{subequations}
\label{eq:double_integrator}
\begin{align}
\min_{\theta} \quad  \mathbb{E}_{x_0 \sim P_{x_0}}&[\Phi(x(T))],\\
\dot{x}_1(t) &= x_2(t), \label{double_integrator_dynamics_1} \\
\dot{x}_2(t) &= u(t), \label{double_integrator_dynamics_2} \\
\label{double_integrator_constraint}
x_2(t) &\leq x_2^{ub} = 2.2,\\
u(t) \in \mathcal{U} &:= [-10,\,10], \label{double_integrator_policy}
\end{align}
\end{subequations}
where $x_1$ is position [m], $x_2$ is velocity [m/s], and $u$ is acceleration [m/s$^2$]. The time horizon is $T=1.5\,\mathrm{s}$ and the target state is $z=[1,\,0]^\top$ (i.e., position $1$ m with zero velocity). We consider two cases: unconstrained L-NODEC wherein the state constraint \eqref{double_integrator_constraint} is ignored and constrained L-NODEC that solves \eqref{eq:double_integrator}. During training, initial states are sampled as $x_1(0)\sim \text{TruncNormal}(0,\,0.025^2, -0.1, 0.1)$ and $x_2(0)=0$
to promote robust convergence from varied initial positions. The potential function is defined as in \eqref{potential_function} with
\[
P \;=\; \begin{bmatrix} 1 & 0 \\[2pt] 0 & 10^{-6} \end{bmatrix},
\]
which places a much higher weight on position error than on velocity error.\footnote{In Algorithm~\ref{algo_lnodec_training}, hyperparameters are set as: $M=200$, $B=3$, $\Gamma=500$, $\alpha=0.02$, $\kappa=5$, $\beta = 15$.}

Fig.~\ref{fig:double_integrator_phase_portraits} shows the phase portraits of state trajectories for L-NODEC, L-NODEC (Constrained), and NODEC. Trajectories corresponding to perturbations in the initial state $x(0)$ are also displayed. The L-NODEC nominal and adversarial trajectories all reach the desired state (position and velocity) in a stable manner; the trajectories never overshoot the desired position and terminate with zero velocity. Since it is uninformed of the constraint \eqref{double_integrator_constraint}, some adversarial trajectories violate it in their paths. As for the L-NODEC constrained policy, the presence of the constraint \eqref{double_integrator_constraint} alters the trajectories to have lower maximum velocity. As a consequence, the trajectories approximately achieve the target state, without terminating exactly at $x_1 = 1$ or achieving zero terminal velocity. Therefore, there is a clear tradeoff between constraint satisfaction and stability, as seen with the unconstrained and constrained L-NODEC policies. However, NODEC yields significantly different trajecotires since the Lyapunov condition is not imposed in the loss function. Inspection of its phase         portrait shows that while the nominal trajectory achieves the target position, the adversarial trajectories do not. Furthermore, the velocities are much lower in magnitude and none of them terminate at zero, which indicates the controlled system will still move after terminal time.

Fig.~\ref{fig:double_integrator_decay} shows the Lyapunov decay \eqref{exponential_stability}, defined as the ratio of the potential functions at $t$ and $t=0$, i.e., $V(x(t))/ V(x(0))$, for the three approaches. The decay is shown over the nominal and 20 advesarial trajectories for each approach. We observe that the L-NODEC policies cannot satsify the Lyapunov condition for the entire time domain $[0,T]$, but the loss formulation minimizes the time spent in violation of it. We recognize that this is not a failure of the proposed method, but the fact that in most real physical systems, the nature of the dynamics restrict the policy's ability to satisfy the Lyapunov condition for the entire time domain. Although the constrained trajectories have a tradeoff with stability, they still satisfy the Lyapunov condition for a comparable period of time as the unconstrained case, highlighting that adding constraints do not offer major drawbacks to performance. Since the NODEC policy is not optimized with respect to the Lyapunov loss, it spends most of the time domain not satisfying the Lyapunov condition, which is the cause for the subpar performance seen in Fig.~\ref{fig:double_integrator_phase_portraits} (right).

\subsection{Control of Thermal Dose Delivery in Plasma Medicine}

Cold atmospheric plasmas (CAPs) are used for treatment of heat-sensitive biomaterials in plasma medicine \citep{laroussi2021low,bonzanini2021perspectives}.   We focus on optimal control of cumulative thermal effects of a biomedical CAP device on a surface. The control objective is to deliver a desired amount of thermal dose, quantified in terms of cumulative equivalent minutes (CEM) \citep{gidon2017effective}, to a surface while maintaining the surface temperature below a safety-critical threshold. The OCP is formulated as \citep{rodrigues2023}
\begin{subequations}
\label{eq:appj}
\begin{align}
\min_{\theta} &\quad  \mathbb{E}_{x_0 \sim P_{x_0}}[\Phi(x(T))],\\
\dot{x}_1(t) &= \frac{u(t)}{3.1981}
               - \frac{0.8088}{\ln\!\big(x_1(t)-25\big) - \ln\!\big(x_1(t)-35\big)},
               \label{appj_dynamics_1}\\
\dot{x}_2(t) &= \frac{0.5^{\,\big(43 - x_1(t)\big)}}{60}, \label{appj_dynamics_2}\\
u(t) &= \pi_\theta\big(x(t)\big) \in [1,\,5], \label{appj_policy}\\
x_1(t) &\le x_1^{\mathrm{ub}}=45. \label{appj_constraint}
\end{align}
\end{subequations}
Here, $x_1$ is surface temperature [$^\circ$C], $x_2$ is thermal dose CEM [min], and $u$ is CAP power [W]. The treatment horizon is $T=100\,\mathrm{s}$. The initial and target states are $x(0)=(37,\,0)$ and $z=(37,\,1.5)$, respectively. The potential is defined as in \eqref{potential_function} with
\[
P=\begin{bmatrix} 10^{-10} & 0 \\[2pt] 0 & 10^{-2} \end{bmatrix},
\]
which emphasizes accuracy in the delivered dose while softly weighting temperature deviation around the target. For this case study, we compare L-NODEC (Constrained) and NODEC.\footnote{In Algorithm~\ref{algo_lnodec_training}, hyperparameters are set as: $M=400$, $B = 4$, $\Gamma=1000$, $\alpha=0.02$, $\kappa=5$, $\beta=15$.}

\begin{figure}
    \centering
    \includegraphics[width=0.98\linewidth]{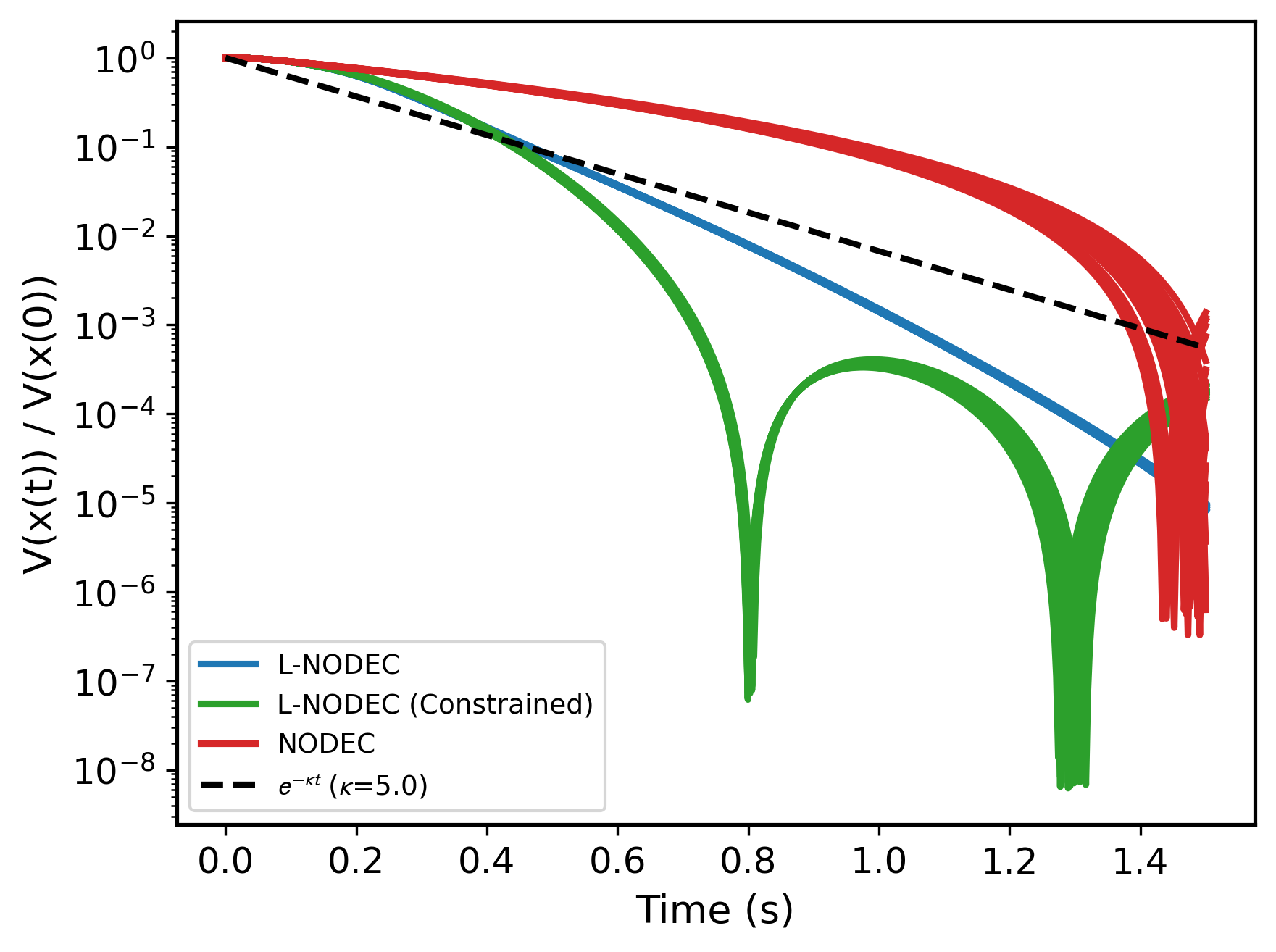}
    \caption{Lyapunov decay, defined as the ratio of the potential functions at times $t$ and $t = 0$, for L-NODEC, L-NODEC (Constrained), and NODEC. The decay is shown for nominal and 20 adversarial trajectories. The exponential stability threshold for $\kappa =5$ is shown in black.}
    \label{fig:double_integrator_decay}
\end{figure}

Fig.~\ref{fig: appj} shows the CEM delivered to the target surface, surface temperature, and control input of the applied power for each strategy. \textcolor{black}{50 adversarial trajectories are generated based on a $2^\circ$C perturbation radius around the nominal initial temperature via uniform sampling.} The state trajectories are truncated when the desired thermal dose of 1.5 min is reached to avoid excessive thermal dose delivery (i.e., plasma treatment is aborted). The average time to reach CEM of 1.5 min is 46 s and 96 s for L-NODEC and NODEC, respectively. This is significant in plasma medicine since shorter treatment times are desirable due to patient safety and comfort \citep{bonzanini2021perspectives}. 
L-NODEC initially maintains the applied power at a higher level (Fig.~\ref{fig: appj}(c)), which results in higher temperature in the initial phase of the treatment, leading to quicker accumulation of CEM till the temperature constraint is reached. Then, L-NODEC reduces the applied power level once it is close to reaching the target CEM. As for NODEC, the lack of information on the target CEM until the final inference time $T$ results in trajectories that take considerably longer to achieve CEM of 1.5 min. This is reflected by the even lower maximum applied power. These results suggest that L-NODEC provides a versatile formulation that aids in reducing the inference time.

\begin{figure*}
    \centering
    \includegraphics[width=1\linewidth]{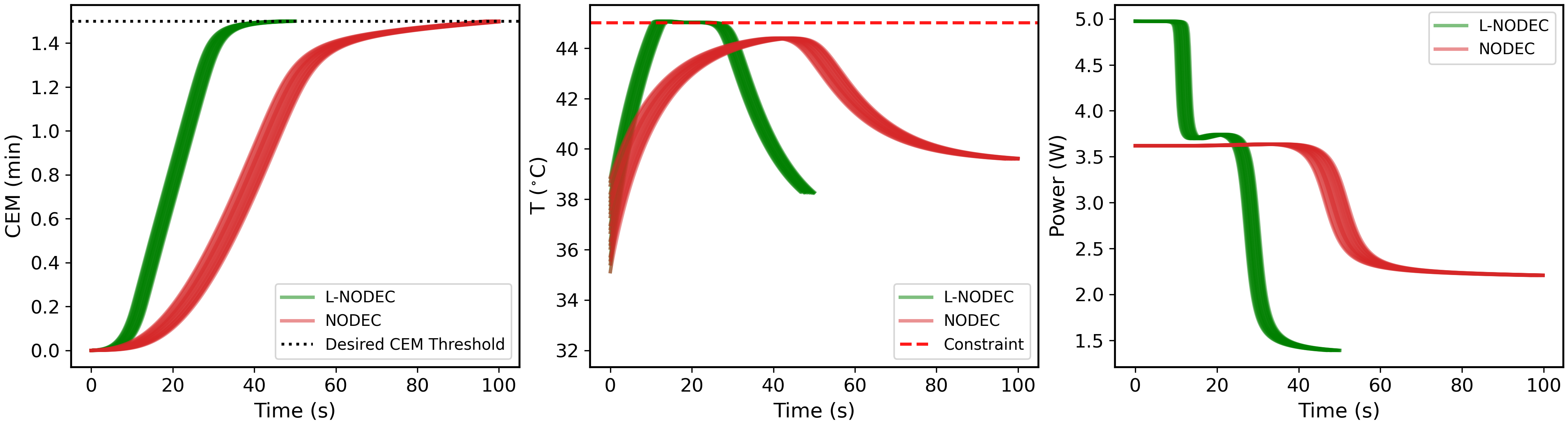}
    \caption{Optimal control of thermal dose delivery of cold atmospheric plasma to a target surface for L-NODEC. (a)  The delivered thermal dose CEM. (b) Surface temperature. (c) Control input, i.e., applied power to plasma.
    50 adversarial trajectories with initial states in a perturbation radius of $2^\circ$C around the nominal initial temperature are simulated and truncated once the desired CEM threshold is met. L-NODEC enables shorter treatment protocols, which are highly desirable in plasma medicine.}
    \label{fig: appj}
\end{figure*}

\section{Conclusion}

This paper investigated the exponential stability and adversarial robustness of the neural ordinary differential equation (NODE) framework for solving continuous-time optimal control problems. The proposed method employs a Lyapunov-based loss formulation that integrates an exponentially stabilizing control Lyapunov function, thereby guaranteeing exponential stability of the controlled system. Numerical results highlighted the importance of enforcing exponential stability in enhancing robustness to perturbations in the initial state. Future work will explore alternative strategies for constraint handling and their impact on stability and performance.



\bibliography{ifacconf}             
                                                   







\appendix
\end{document}